\documentclass[letterpaper, 10 pt, conference]{ieeeconf}  

\IEEEoverridecommandlockouts                              

\usepackage{graphics} 
\usepackage{amsmath} 
\usepackage{amssymb}  
\usepackage{graphicx}
\usepackage{xcolor}
\usepackage{amsmath}  
\usepackage{amssymb}  
\usepackage{bm}      

\usepackage[capitalise]{cleveref}
\usepackage{booktabs}
\usepackage{array}
\usepackage{multirow} 
\usepackage{makecell}
\usepackage{siunitx}
\usepackage{fancyhdr}

\title{\LARGE \bf
ProTrav: Probabilistic Traversability with Elevation Uncertainty Propagation for Risk-aware Off-road Navigation}

\author{Chuanlin Zuo$^{1}$, Milo\v{s} Pr\'{a}gr$^{1}$, Arun Kumar Singh$^{2}$, Mohammad Loni$^{3}$ and Reza Ghabcheloo$^{1}$  
\thanks{*This work was supported by the European Union’s Horizon Europe Marie Skłodowska-Curie Actions Doctoral Networks (MSCA-DN-ID) HORIZON-MSCA-2024-DN call, under the Grant Agreement number 101226234, and was in part supported by Horizon Europe Project XSCAVE under Grant 101189836.
}%
\thanks{$^{1}$ are with Faculty of Engineering and Natural Sciences, Tampere University, Finland.
        {\tt\small chuanlin.zuo@tuni.fi, milos.pragr@tuni.fi, reza.ghabcheloo@tuni.fi}}%
\thanks{$^{2}$ is with the Institute of Technology, University of Tartu, Estonia.
        {\tt\small arun.singh@ut.ee}}%
\thanks{$^{3}$ is with the AI and ML Department, Volvo Autonomous Solutions, Sweden.
        {\tt\small mohammad.loni@volvo.com}}%
}

\begin{document}

\maketitle

\begin{abstract}
Reliable traversability estimation is essential for autonomous navigation on unstructured terrain.
State-of-the-art approaches typically evaluate traversability from deterministic terrain elevation, thereby overlooking spatial correlations in the terrain and failing to account for how elevation uncertainty propagates through nonlinear traversability mappings.
We propose ProTrav, a probabilistic terrain representation that uses function-space sampling to generate spatially correlated terrain realizations and propagate elevation uncertainty to traversability cost.
First, a Sparse Variational Gaussian Process (SVGP) reconstructs a probabilistic elevation field from observations. 
Then, posterior function-space sampling generates spatially correlated terrain realizations, preserving spatial dependencies between neighboring locations while propagating elevation uncertainty through geometric features into traversability cost.
Consequently, a distribution of traversability cost at arbitrary location is established, where the mean and standard deviation quantify the expected traversability cost and associated uncertainty, respectively.
A risk term is defined by moment-based Upper Confidence Bound (UCB) and incorporated into a Model Predictive Path Integral (MPPI) controller for risk-aware navigation.
Simulation experiments demonstrate navigation success rates exceeding 90\% on different terrains, compared with below 40\% for a baseline, while reducing attitude variations.
Furthermore, the proposed approach is validated on a robot traversing real-world off-road terrain.
\end{abstract}

\newcommand{\obs}{\ensuremath{\boldsymbol{p}}}
\newcommand{\obsmat}{\ensuremath{\boldsymbol{P}}}
\newcommand{\pos}{\ensuremath{\boldsymbol{x}}}
\newcommand{\posvec}{\ensuremath{\boldsymbol{X}}}
\newcommand{\elev}{\ensuremath{z}}
\newcommand{\elevvec}{\ensuremath{\boldsymbol{\elev}}}

\newcommand{\meanf}{\ensuremath{m}}
\newcommand{\kernelf}{\ensuremath{k}}
\newcommand{\rbflscale}{\ensuremath{\ell_{s}}}
\newcommand{\rbfsigma}{\ensuremath{\sigma_{s}}}
\newcommand{\noisesigma}{\ensuremath{\sigma_{n}}}
\newcommand{\covmat}{\ensuremath{\boldsymbol{K}}}
\newcommand{\covmatcross}{\ensuremath{\boldsymbol{\kernelf}}}

\newcommand{\normal}{\ensuremath{\boldsymbol{n}}}
\newcommand{\slope}{\ensuremath{{s}}}
\newcommand{\roughness}{\ensuremath{{r}}}
\newcommand{\steph}{\ensuremath{{h}}}
\newcommand{\geom}{\ensuremath{\boldsymbol{g}}}
\newcommand{\geomvec}{\ensuremath{\boldsymbol{G}}}

\newcommand{\criticalthr}{\ensuremath{\tau}}
\newcommand{\critical}{\ensuremath{\boldsymbol{\tau}}}
\newcommand{\cost}{\ensuremath{c}}
\newcommand{\costvec}{\ensuremath{\boldsymbol{\cost}}}
\newcommand{\score}{\ensuremath{\cost_{\geom}}}

\newcommand{\costucb}{\ensuremath{\cost_\text{UCB}}}
\newcommand{\factorconf}{\ensuremath{\kappa}}
\newcommand{\costmppi}{\ensuremath{\rho}}
\newcommand{\horizon}{\ensuremath{H}}

\section{INTRODUCTION}

Autonomous navigation in off-road environments depends on an accurate representation of terrain traversability \cite{papadakis2013terrain,beycimen2023comprehensive}.
Unlike structured roads, off-road terrain exhibits substantial geometric variation, where slope, roughness, and step height directly affect robot stability and mobility.
We focus on LiDAR, which is widely used for terrain perception due to its ability to provide accurate distance measurements. 
However, in practical environments, LiDAR observations are inherently sparse, irregularly distributed, and affected by measurement noise and occlusion.
Therefore, uncertainty is an intrinsic property of terrain perception.

\begin{figure}[t]
    \centering
    \includegraphics[width=1\linewidth]{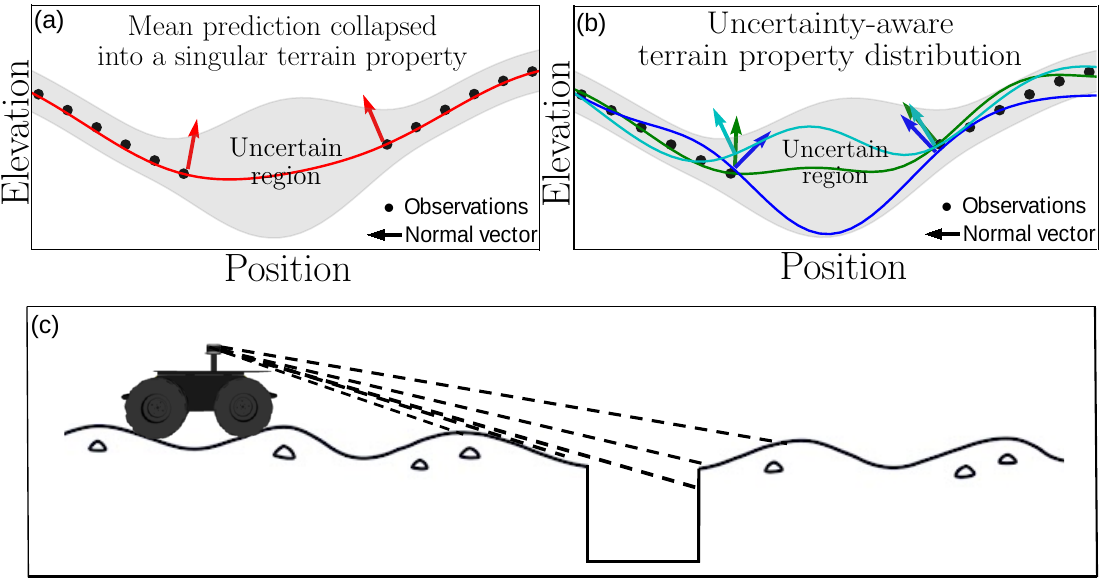}
    \caption{%
        Comparison of the mean-first strategy (a) and proposed uncertainty-aware method (b), represented as terrain normals. Within a negative obstacle (c), the terrains observations are missing, and hence the terrain elevation prediction is uncertain.
        (a) The mean prediction of terrain elevation  defines a single terrain property per query position. Since the mean elevation prediction remains flat despite higher uncertainty, the terrain descriptors are overly optimistic, indicating flat and thus safe terrain. 
        (b) Multiple samples of terrains surface define multiple terrain properties per query position. Therefore, the elevation uncertainty propagates into the geometric description, highlighting possibly unsafe uncertain region.
      }
    \label{fig:normal_feature_sampling}
\end{figure}

Recent work has begun incorporating probabilistic terrain models into traversability analysis and path planning to account for this perceptual uncertainty \cite{fan2021step}.
In most such approaches, however, geometric features are computed from a single deterministic elevation estimate, typically the posterior mean of a probabilistic terrain elevation model \cite{leininger2024gaussian}.
Approaches that explicitly enumerate elevation uncertainty commonly use it only as a mask or inject it into the cost through heuristic inflation, rather than propagating it through subsequent feature extraction \cite{sevastopoulos2022survey}.
Such mean-first strategy introduces a fundamental limitation for traversability estimation because geometric features are usually nonlinear mappings of elevation.
By Jensen's inequality, the expected feature generally cannot be recovered by evaluating a nonlinear mapping at the expected elevation.
This issue can arise, for example, near a partially observed negative obstacle as shown in \cref{fig:normal_feature_sampling}.
There, mean elevation field may smooth the underlying discontinuity and underestimate its risk.
More generally, extracting features directly from the expected elevation can suppress terrain variations that are critical for identifying steep slopes, steps, and partially observed regions.
Furthermore, treating elevation uncertainty independently at individual cells neglects the spatial correlations of continuous terrain.
Since geometric features are inherently neighborhood operators instead of one location, their uncertainty depends on the joint uncertainty of the underlying elevation patch.
Ignoring these spatial dependencies can therefore further prevent faithful characterization of traversability \cite{gonga2011techniques}.

We address these limitations by formulating terrain traversability estimation as a probabilistic terrain representation problem.
The terrain surface is modeled as a continuous stochastic function using a Sparse Variational Gaussian Process (SVGP).
We draw function-space realizations from the SVGP posterior, obtaining spatially correlated elevation fields. 
For each realization, slope, roughness, and step height are computed and combined into a traversability cost, from which an empirical distribution of traversability costs is obtained across the realizations.
The mean and standard deviation of this distribution characterize the expected traversability cost and its associated uncertainty, respectively.
Based on these moments, a risk term is constructed using an Upper Confidence Bound (UCB) and incorporated into the cost function of a Model Predictive Path Integral (MPPI) controller, forming a risk-aware navigation framework.
The main contributions of this work are summarized as follows.
\begin{itemize}
\item Terrain traversability estimation is formulated as a probabilistic terrain representation problem, explicitly propagating perception and elevation uncertainty to traversability cost.
\item A function-space sampling strategy based on the posterior used to generate spatially correlated terrain realizations and characterize uncertainty in nonlinear geometric and traversability mapping. 
\item A risk-aware MPPI navigation framework integrating ProTrav with UCB-based risk, validated across diverse terrains in both simulation and real-world experiments.
\end{itemize}
The remainder of this paper is organized as follows.
Section II reviews related work on terrain traversability estimation.
Section III introduces preliminary methods and concepts. 
Section IV presents the proposed probabilistic terrain traversability representation and risk-aware navigation framework.
Section V describes the simulation and real-world experiments.
Section VI concludes the paper.

\section{RELATED WORK}

\subsection{Terrain Traversability Estimation}
Traditional traversability estimation characterizes terrain using handcrafted, human interpretable descriptors of terrain surface geometry, such as slope, roughness, and step height~\cite{tanaka2015fuzzy,meng2018terrain}.
Such descriptors can be computed over grid maps, or directly assess terrain geometry from point clouds, omitting explicit surface reconstruction~\cite{krusi2017driving}.
Alternatively, learning-based methods infer traversability directly from terrain observations, reducing reliance on manually designed geometric features. 
For instance, Chavez-Garcia et al. \cite{chavez2018learning} learn ground traversability from height maps using a convolutional neural network, while Wu et al. \cite{wu2025real} convert sparse point clouds into dense, vehicle-centered bird's-eye-view traversability maps through efficient voxelization and learned completion. 

Recent approaches extend learning-based traversability estimation to multimodal and long-range perception. 
RoadRunner \cite{frey2024roadrunner} predicts terrain traversability and elevation from multi-camera and LiDAR observations, and its multi-range, multi-resolution extension~\cite{patel2024roadrunner} estimates traversability and elevation at different spatial scales to support long-range off-road navigation. 
METAVerse \cite{seo2024metaverse} employs meta-learning to adapt a generalized traversability model to previously unseen off-road environments using recent vehicle–terrain interaction data. 
Other approaches exploit multi-modal self-supervision~\cite{fang2025scene} or implicit neural representations~\cite{jia2026off} to jointly encode geometric and semantic terrain information for continuous traversability estimation and terrain-aware planning. 
These approaches provide increasingly expressive traversability representations, but generally produce a deterministic traversability from terrain observations.

\subsection{Probabilistic Traversability}
Terrain perception is inherently uncertain due to sensor noise, sparse data, localization errors, and uncertain vehicle–terrain interactions. 
Probabilistic terrain mapping represents uncertainty directly in the terrain model. 
Fankhauser et al. \cite{fankhauser2018probabilistic} incorporate localization and range-sensor uncertainty into elevation mapping, and Jung et al. \cite{jung2025uncertainty} model sharp terrain structures and provide predictive uncertainty together with the estimated geometry using Neural Processes. 
Kolu et al. \cite{kolu2015mapping} address the effects of localization and sensor calibration errors on terrain perception reliability through segment-based processing of laser measurements, improving the robustness of obstacle mapping.
Hansen and Wettergreen \cite{hansen2023range} propose a range-based Gaussian Process (GP) mapping framework that models LiDAR range as a function of spherical sampling angles. 
Others model spatially correlated terrain uncertainty~\cite{raja2025proterrain} and propagate uncertainty arising from vehicle–terrain interactions into predicted motion~\cite{lee2023learning}. 

Alternatively, uncertainty can be modeled directly at the traversability and risk levels. 
Endo et al. \cite{endo2026deep} learn wheel-slip distributions and decomposed predictive uncertainty into aleatoric and epistemic components. 
STEP \cite{fan2021step} uses probabilistic terrain risk and Conditional Value-at-Risk (CVaR) for conservative traversability estimation, while Velociraptor~\cite{triest2024velociraptor} learns cost, speed, and uncertainty maps from multimodal observations and uses CVaR for risk-aware navigation. 

Closest to our work are the recent works that combine probabilistic terrain models with geometric traversability estimation. 
Li et al. \cite{li2026real} estimate slope, roughness, and step height from temporally fused elevation maps for real-time traversability mapping. 
Tan et al. \cite{tan2025real} incorporate geometric features into a sparse GP and use predictive uncertainty to adapt the confidence of historical traversability estimates. 
Leininger et al. \cite{leininger2024gaussian} use a GP-based terrain representation to estimate geometric traversability and employ elevation uncertainty to mask poorly observed regions. 
However, these approaches do not explicitly characterize uncertainty of the traversability cost, which is a nonlinear mapping of observations and elevation.
As a result, uncertainty is not explicitly propagated to the traversability level.
In this work, we address this gap through function-space sampling, propagating elevation uncertainty through nonlinear mapping to obtain a probabilistic traversability cost representation, and further construct a risk term for risk-aware navigation.


\begin{figure*}[t]
    \centering
    \includegraphics[width=1\linewidth]{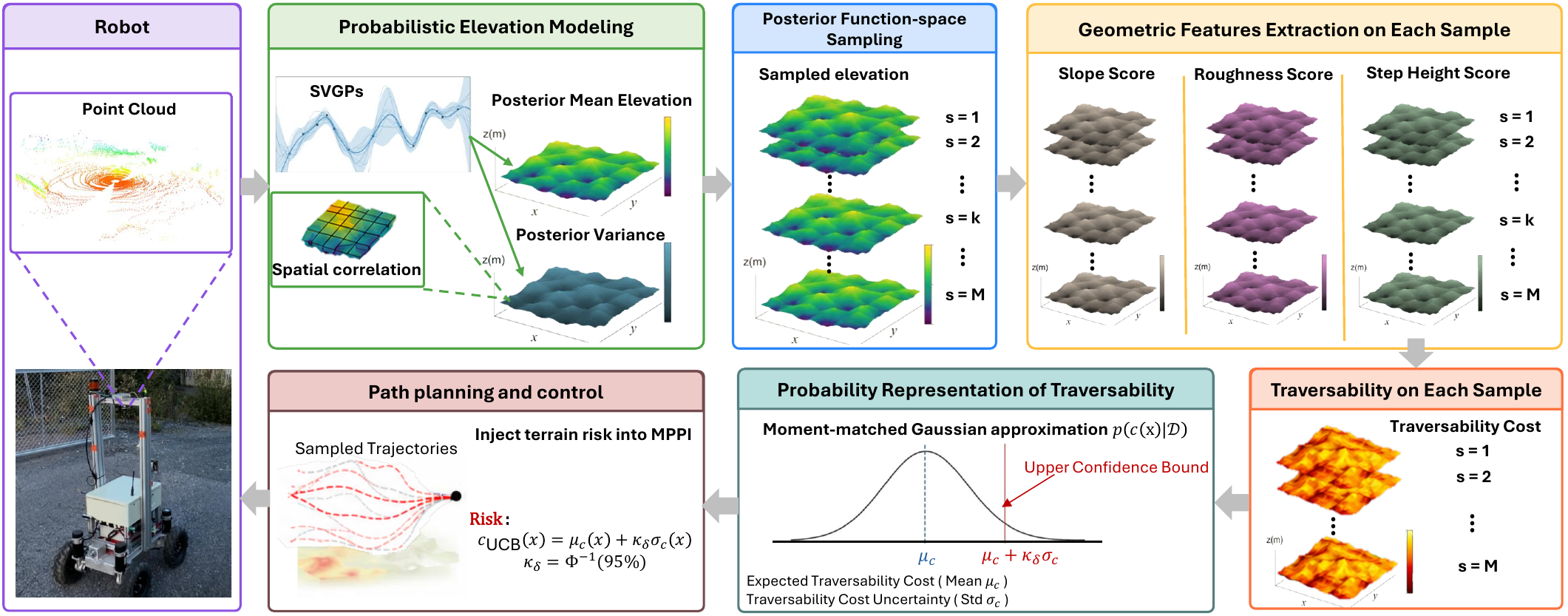}
    \caption{Overview of the proposed probabilistic terrain traversability representation and navigation framework. A point cloud is used to train an SVGP to construct a continuous probabilistic elevation field. Terrain realizations are generated through function-space sampling and terrain geometric features are extracted from each realization and fused into traversability costs. These realizations are aggregated to obtain moment-based Gaussian approximation, from which a risk term is constructed by UCB and injected into the MPPI rollout cost. The resulting velocity command is sent to the robot for execution.}
    \label{fig:system_design}
    \vspace{-1.25em}
\end{figure*}

\section{PRELIMINARIES}
\label{sec:preliminaries}

\subsection{Gaussian Process Terrain Modeling}
\label{subsec:GP}
A terrain surface can be represented as a continuous elevation field inferred from a set of spatial observations. 
Let $\obs_i=(\pos_i,\elev_i)$ denote an observation, where $\pos_i=\left[x_i,y_i\right]^\top$ is the horizontal location and $z_i$ is the measured terrain elevation. The observation model is given by
\begin{equation}
\elev_i=f(\pos_i)+\epsilon_i,
\qquad
\epsilon_i\sim\mathcal{N}(0,\noisesigma^2),
\end{equation}
where $f:\mathbb{R}^2 \rightarrow \mathbb{R}$ denotes the terrain elevation function and $\sigma_n^2$ is the measurement noise variance. The function $f$ can be modeled as a Gaussian Process \cite{schulz2018tutorial}
\begin{equation}
\begin{gathered}
f(\pos)\sim \mathcal{GP} \left(\meanf(\pos),\kernelf(\pos,\pos')\right),\\
\kernelf(\pos, \pos') = \rbfsigma^2 \exp \left( -\frac{\|\pos - \pos'\|^2}{2\rbflscale^2} \right),
\end{gathered}
\end{equation}
where $\meanf(\pos)$ and $\kernelf(\pos,\pos')$ are the prior mean function and covariance kernel.
We use a Radial Basis Function (RBF) kernel with output scale $\rbfsigma^2$ and length scale $\rbflscale$.

Given the observation set $\mathcal{D}=\left\{\obs_i\right\}_{i=1}^{N_o}$, where $N_o$ is the number of observations, the GP induces a posterior distribution of the terrain elevation field. For the query location $\pos^*$, the posterior elevation distribution is given by
\begin{equation} 
\label{eq:posterior_gp}
\begin{gathered}
p(f(\pos^*) \mid \mathcal{D}) \sim \mathcal{N}\left(\mu_f(\pos^*), \sigma_f^2(\pos^*)\right),\\
\mu_f(\pos^*) = \covmatcross_*^\top \left( \covmat + \noisesigma^2 \mathbf{I} \right)^{-1} \elevvec,\\
\sigma_f^2(\pos^*) = \kernelf(\pos^*, \pos^*) - \covmatcross_*^\top \left( \covmat + \noisesigma^2 \mathbf{I} \right)^{-1} \mathbf{\kernelf}_*,
\end{gathered}
\end{equation}
where $\elevvec = [\elev_1, \dots, \elev_{N_o}]^\top$, $\covmat \in \mathbb{R}^{N_o \times N_o}$ with $\covmat_{ij} = \kernelf(\pos_i, \pos_j)$ is the covariance matrix of the observed points, and $\covmatcross_* = [\kernelf(\pos^*, \pos_1), \dots, \kernelf(\pos^*, \pos_{N_o})]^\top$ the cross-covariance vector between the query location and the observations.

Exact GP inference incurs $\mathcal{O}(N_o^3)$ complexity that is impractical for online terrain mapping. 
We consequently  employ the SVGP \cite{hensman2015scalable}, which approximates the GP posterior using a small set of inducing points, reducing inference cost to a level suitable for real-time mapping. 

\subsection{Terrain Geometric Features Extraction}
\label{geo feature}

We adopt the terrain feature descriptors from~\cite{li2026real,jian2022putn,sade2026adaptive}.
Slope describes the inclination of the local terrain surface and is primarily associated with vehicle stability and rollover risk.
Let a window containt $N_w$ terrain points $\left\{\obs_j\right\}_{j=1}^{N_w}$.
Its centered point matrix is $\obsmat = [\obs_1 - \bar{\obs}, \dots, \obs_{N_w} - \bar{\obs}]^\top \in \mathbb{R}^{N_w \times 3}$, where $\bar{\obs}$ is the window mean point.
The dominant terrain plane is estimated using Singular Value Decomposition (SVD)
\begin{equation}
\begin{gathered}
\obsmat = \mathbf{U}\mathbf{\Sigma}\mathbf{V}^{\top}, \\
\mathbf{\Sigma} = \operatorname{diag}(\sigma_1,\sigma_2,\sigma_3), \quad
\mathbf{V} = [\mathbf{v}_1, \mathbf{v}_2, \mathbf{v}_3],
\end{gathered}
\end{equation}
where $\sigma_1 \geq \sigma_2 \geq \sigma_3 \geq 0$, 
and the window terrain normal $\normal$ is given by $\normal = \mathbf{v}_3$.
Let the $\elev$-axis be aligned with gravity, $\mathbf{e}_z = [0,0,1]^\top$. 
Then, the window slope is the angle between the terrain normal and $\mathbf{e}_z$~\cite{jian2022putn},
\begin{equation}
\slope = \arccos\left(|\normal^{\top}\mathbf{e}_z|\right).
\end{equation}

Roughness quantifies window terrain deviations from the dominant terrain plane, thus capturing small-scale irregularities that induce vibration and unstable ground contact.
In \cite{sade2026adaptive}, roughness is defined as the root-mean-square orthogonal distance of the points from the best-fitting plane
\begin{equation}
\roughness =
\sqrt{\frac{1}{N_w} \sum_{j=1}^{N_w} d_j^2},
\end{equation}
where $d_j$ is the orthogonal distance from $\obs_j$ to the plane. 

Finally, step height characterizes abrupt window elevation changes and is primarily associated with obstacle and potential collision.
Following \cite{li2026real}, step height measures the maximum elevation variation within the window
\begin{equation}
\steph = \max_{i} \elev_i - \min_{j} \elev_j, \qquad i,j \in \{1,\dots,N_w\}.
\end{equation}

\section{METHODOLOGY}
ProTrav aims to improve safety of mobile robots driving in uncertain, unstructured environments by propagating elevation uncertainty to geometric descriptors and traversal costs as overviewed in~\cref{fig:system_design}.
First, a probabilistic terrain elevation is built from terrain observations using SVGP.
Next, elevation maps are drawn from the GP posterior function-space. 
Each map is used to independently compute geometric descriptors and traversability costs, which are then used to construct resulting traversability cost distribution map.
Finally, we show how this map can be used for risk-aware navigation, selecting to use an MPPI with UCB-based risk.

\subsection{Function-Space Elevation Map Sampling}
\label{subsec:sampling}
From \cref{subsec:GP}, the posterior distribution of the elevation $p(f(\pos^*) \mid \mathcal{D})$ can be obtained at a query location $\pos^*$.
Let any arbitrary traversability cost $\cost$ be a nonlinear mapping of observations and elevation field.
Then, evaluating this mapping only at the posterior mean $\mu_f(\pos^*)$ neglects the uncertainty in the inferred elevation field, as illustrated in \cref{fig:normal_feature_sampling}.
In general, due to the nonlinear nature of the mapping, the traversability cost evaluated at the elevation mean is not equal to the expected traversability cost, $\cost(\mathbb{E}[f]) \neq \mathbb{E}[\cost(f)]$, by Jensen's inequality.
This discrepancy implies that computing traversability solely from the posterior mean can provide an inaccurate approximation of the expected traversability cost and fail to capture terrain variations induced by elevation uncertainty.

The resulting distribution of traversability cost is characterized by propagating the joint posterior of the terrain elevation field through the geometric feature extraction and  traversability cost computation using Monte Carlo sampling in function space. 
We discretize the local map surrounding the robot into $N$ grid locations $\posvec^*=\left[\pos_1^*,\pos_2^*,\dots,\pos_N^*\right]^\top$ in the vehicle-relative frame with a fixed spatial offset from the vehicle-center point.
Given the GP posterior~\eqref{eq:posterior_gp}, the joint elevation field distribution over the local map is
\begin{equation}
\begin{gathered}
p(f(\posvec^*) \mid \mathcal{D})
\sim
\mathcal{N}
\left(
\boldsymbol{\mu}_*,
\mathbf{\Sigma}_*
\right), 
\\
\boldsymbol{\mu}_*
=
\left[
\mu_f(\pos_1^*),
\mu_f(\pos_2^*),
\dots,
\mu_f(\pos_N^*)
\right]^\top, 
\\
\left[\mathbf{\Sigma}_*\right]_{ij}
=
\kernelf(\pos_i^*,\pos_j^*)
-
\covmatcross_{i,*}^{\top}
\left(
\covmat+\noisesigma^2\mathbf{I}
\right)^{-1}
\covmatcross_{j,*},
i,j=1,\dots,N.
\end{gathered}
\end{equation}

We draw $M$ realizations from this joint posterior, $\mathcal{F}=\left\{f^{(1)}(\posvec^*),f^{(2)}(\posvec^*), \dots,f^{(M)}(\posvec^*)\right\}$, where each realization represents a plausible terrain surface consistent with the observations and the posterior uncertainty.
Rather than sampling each location independently from its marginal posterior distribution, we jointly sample the local elevation map using the full posterior covariance $\Sigma_*$.
This preserves the spatial correlations induced by the GP within each realization.

\subsection{Geometry-Based Traversability Mapping}
\label{meth-traversability}
Given a terrain elevation field, we use geometric features in \cref{geo feature} to define traversability.
The geometric descriptor is given by
\begin{equation}
  \geom(\pos^*) = [\slope(\pos^*),\roughness(\pos^*),\steph(\pos^*)]^\top.
\end{equation}

Each feature is normalized by a critical threshold $\critical = [\criticalthr_\slope, \criticalthr_\roughness, \criticalthr_\steph]^\top$ which is selected according to robot's geometric and dynamic capabilities. 
The resulting weighted geometric score is defined as
\begin{equation}
\score(\pos^*) = \mathbf{w}^\top \operatorname{diag}(\critical)^{-1} \geom(\pos^*),
\end{equation}
where $\mathbf{w} = [w_\slope, w_\roughness, w_\steph]^\top$ specifies the relative contribution of each feature, subject to $\mathbf{w} \succeq \mathbf{0}$ and $w_\slope + w_\roughness + w_\steph = 1$.
A Hill-type function then bounds and smoothens the score as
\begin{equation}
\cost(\pos^*)
=
\frac{2\score(\pos^*)^\gamma}
  {1+\score(\pos^*)^\gamma},
\end{equation}
where $\gamma > 0$ controls the sharpness of the transition between low- and high-cost terrain.
We set $\gamma=3$ empirically.
The transformation suppresses the influence of small geometric variations in benign terrain while preventing extreme feature values from disproportionately dominating the traversability cost.

For each realization from \cref{subsec:sampling}, the corresponding geometric features and traversability cost are computed.
Repeating this process over all $M$ terrain realizations yields a set of traversability cost realizations, $\costvec=\left\{\cost^{(1)}(\posvec^*),\cost^{(2)}(\posvec^*), \dots,\cost^{(M)}(\posvec^*)\right\}$.
These realizations capture how the uncertainty in the elevation field propagates through the nonlinear geometric feature extraction and traversability cost computation. We therefore represent traversability probabilistically by estimating the distribution of the traversability cost from these realizations,
\begin{equation}
p(\cost(\posvec^*)\mid\mathcal{D})
\approx
\frac{1}{M}
\sum_{m=1}^{M}
\delta
\left(
\cost(\posvec^*)
-
\cost^{(m)}(\posvec^*)
\right),
\end{equation}
where $\delta(\cdot)$ is Dirac delta function, constituting the proposed probabilistic representation of terrain traversability cost.

\subsection{Risk-Aware MPPI Navigation}
We integrate the proposed probabilistic traversability cost into a MPPI controller to account jointly for terrain difficulty and uncertainty during trajectory evaluation.
Since the traversability cost distribution obtained through nonlinear mapping is generally non-Gaussian, we construct a uncertainty-aware risk term using its first two moments. Specifically, we define a moment-based UCB risk term, 
\begin{equation}
\begin{gathered}
\costucb(\pos)
=
\mu_\cost
\left(
\pos
\right)
+
\factorconf_{\delta}
\sigma_\cost
\left(
\pos
\right),
\quad
\factorconf_\delta =
\Phi^{-1}(1-\delta),
\\
\mu_\cost(\pos)
=
\frac{1}{M}
\sum_{m=1}^{M}
\cost^{(m)}(\pos),
\\
\sigma_\cost(\pos)
=
\sqrt{
\frac{1}{M-1}
\sum_{m=1}^{M}
\left(
\cost^{(m)}(\pos)
-
\mu_\cost(\pos)
\right)^2
},
\end{gathered}
\end{equation}
with $\Phi^{-1}(\cdot)$ denoting the inverse cumulative distribution function of the standard normal distribution and $1-\delta$ specifying the desired one-sided confidence level.
In this work, we use confidence level of $95\%$ yielding $\factorconf_{\delta}=1.645$.

The final MPPI trajectory cost is defined as
\begin{equation}
\begin{gathered}
\costmppi = \costmppi_{\mathrm{goal}}
+
\lambda_r \costmppi_{\mathrm{risk}},
\\
\costmppi_{\mathrm{goal}}
=
\lambda_H\left\|
  \pos_\horizon-\pos_\text{goal}
\right\|_2 
+
\lambda_h \sum_{t=0}^{\horizon-1} 
\left\|
  \pos_t-\pos_\text{goal}
\right\|_2,
\\
\costmppi_{\mathrm{risk}}
=
\sum_{t=0}^\horizon
\costucb
\left(
\pos_t
\right),
\end{gathered}
\end{equation}
where $\pos_t$ is the horizontal location at time step $t$, $\horizon$ is the rollout horizon, $\pos_\horizon$ is the terminal location and $\pos_\text{goal}$ is the goal location. 
$\lambda_H, \lambda_h, \lambda_r > 0$ balance goal-reaching performance against terrain risk. 
By evaluating each MPPI rollout using $\costucb$, the controller jointly accounts for the expected terrain difficulty and the uncertainty of cost estimation.

\section{EXPERIMENT AND RESULTS} 
To experimentally verify the proposed approach, we first carry out a quantitative and qualitative evaluation against a baseline deterministic traversability using a Husky robot in a Gazebo simulation.
Therein, we also carry out ablation tests on uncertainty sensitivity in planning. 
In the rest of the presented results, we report on deployments onboard a small differential drive robot.

\subsection{Simulation Experiments}

\begin{table*}[t]
\centering
\caption{%
Simulated Performance on Mountain Terrain, $\mu\pm\sigma$ of $10$ Trials and Success Rate.
}
\label{tab:sim_result_tab}
\resizebox{\linewidth}{!}{
\begin{tabular}{lll cccc c}
\toprule
\textbf{Start} (m) & \textbf{Goal} (m) & \textbf{Method} & 
\makecell{$|\phi|_{\text{mean}}$ (deg)} & 
\makecell{$|\phi|_{\text{max}}$ (deg)} & 
\makecell{$|\theta|_{\text{mean}}$ (deg)} & 
\makecell{$|\theta|_{\text{max}}$ (deg)} &
\makecell{\textbf{Success Rate}} \\
\midrule

  \multirow{5}{*}{(20, 13)} & \multirow{5}{*}{(7, -1)} 
    & Base-RRT*  & -- & -- & -- & -- & 0.00 \\
  & & Base-MPPI  & 7.74 $\pm$ 4.10 & 21.66 $\pm$ 9.24 & 12.39 $\pm$ 1.45 & 36.66 $\pm$ 1.58 & 100.00\% \\
  & & Prob-MPPI($\factorconf_{\delta}=0$)  & 4.40 $\pm$ 0.11 & 11.87 $\pm$ 0.70 & 3.84 $\pm$ 0.03 & 10.88 $\pm$ 0.12 & 100.00\% \\
  & & Prob-MPPI($\factorconf_{\delta}=0.842$)  & 4.27 $\pm$ 0.08 & 11.16 $\pm$ 0.39 & 3.81 $\pm$ 0.06 & 10.89 $\pm$ 0.18 & 100.00\% \\
  & & Prob-MPPI($\factorconf_{\delta}=1.645$)  & \textbf{4.07 $\pm$ 0.09} & \textbf{10.60 $\pm$ 0.31}& \textbf{3.79 $\pm$ 0.03} & \textbf{10.87 $\pm$ 0.09} & \textbf{100.00\%} \\
  
\cmidrule(lr){1-8}
  \multirow{5}{*}{(-7, -7)} & \multirow{5}{*}{(15, 18)} 
  & Base-RRT*  & 7.21 $\pm$ 3.03 & 21.87 $\pm$ 8.53 & 6.16 $\pm$ 2.42 & 23.52 $\pm$ 8.24 & 30.00\% \\
  & & Base-MPPI  & 4.13 $\pm$ 0.34 & 13.19 $\pm$ 1.14 & 3.58 $\pm$ 0.07 & 10.97 $\pm$ 0.13 & 100.00\% \\
  & & Prob-MPPI($\factorconf_{\delta}=0$)  & 3.84 $\pm$ 0.07 & 12.35 $\pm$ 0.60 & 3.67 $\pm$ 0.04 & 10.22 $\pm$ 0.34 & 100.00\% \\
  & & Prob-MPPI($\factorconf_{\delta}=0.842$)  & 3.65 $\pm$ 0.04 & 12.58 $\pm$ 0.59 & 3.72 $\pm$ 0.03 & 10.14 $\pm$ 0.60 & 100.00\% \\
  & & Prob-MPPI($\factorconf_{\delta}=1.645$)  & \textbf{3.50 $\pm$ 0.09} & \textbf{12.39 $\pm$ 0.89}& \textbf{3.70 $\pm$ 0.04} & \textbf{10.49 $\pm$ 0.82} & \textbf{100.00\%} \\
  
\cmidrule(lr){1-8}
  \multirow{5}{*}{(17, -17)} & \multirow{5}{*}{(-7, -7)} 
  & Base-RRT*  & 8.48 $\pm$ 1.63 & 31.94 $\pm$ 6.59 & 8.31 $\pm$ 2.28 & 33.02 $\pm$ 9.66 & 40.00\% \\
  & & Base-MPPI  & 4.32 $\pm$ 0.63 & 9.53 $\pm$ 0.35 & 4.08 $\pm$ 0.13 & 9.14 $\pm$ 0.39 & 100.00\% \\
  & & Prob-MPPI($\factorconf_{\delta}=0$)  & 2.56 $\pm$ 0.08 & 5.94 $\pm$ 0.40 & 4.42 $\pm$ 0.03 & 10.35 $\pm$ 0.17 & 100.00\% \\
  & & Prob-MPPI($\factorconf_{\delta}=0.842$)  & 2.62 $\pm$ 0.06 & 6.75 $\pm$ 0.59 & 4.37 $\pm$ 0.03 & 10.40 $\pm$ 0.11 & 100.00\% \\
  & & Prob-MPPI($\factorconf_{\delta}=1.645$)  & \textbf{2.52 $\pm$ 0.04} & \textbf{7.34 $\pm$ 0.53}& \textbf{4.33 $\pm$ 0.05} & \textbf{10.39 $\pm$ 0.12} & \textbf{100.00\%} \\

\cmidrule(lr){1-8}
  \multirow{5}{*}{(17, -17)} & \multirow{5}{*}{(-8, 18)} 
  & Base-RRT*  & 7.80 $\pm$ 0.00 & 33.50 $\pm$ 0.00 & 7.60 $\pm$ 0.00 & 31.09 $\pm$ 0.00 & 10.00\% \\
  & & Base-MPPI  & 5.16 $\pm$ 0.10 & 14.82 $\pm$ 0.51 & 7.21 $\pm$ 0.08 & 17.47 $\pm$ 0.06 & 100.00\% \\
  & & Prob-MPPI($\factorconf_{\delta}=0$)  & 3.46 $\pm$ 0.35 & 11.42 $\pm$ 0.51 & 6.97 $\pm$ 1.28 & 14.75 $\pm$ 2.12 & 100.00\% \\
  & & Prob-MPPI($\factorconf_{\delta}=0.842$)  & 3.07 $\pm$ 0.07 & 11.63 $\pm$ 0.60 & 5.93 $\pm$ 0.04 & 13.07 $\pm$ 0.22 & 100.00\% \\
  & & Prob-MPPI($\factorconf_{\delta}=1.645$)  & \textbf{3.04 $\pm$ 0.07} & \textbf{11.73 $\pm$ 0.69}& \textbf{5.88 $\pm$ 0.04} & \textbf{13.10 $\pm$ 0.35} & \textbf{100.00\%} \\

\bottomrule
\end{tabular}}
\vspace{4pt}
\noindent
\begin{minipage}{\linewidth}
\footnotesize
\textit{Note:} Roll and pitch are recored only from successful navigation trials to avoid confounding attitude evaluation with failure modes such as timeout, rollover, or falling. The success rate is computed over all trials. ``--'' indicates that no successful trial was recorded for the corresponding configuration.
\end{minipage}
\vspace{-1.5em} 
\end{table*}

\begin{figure}[t]
    \centering
    \includegraphics[width=1\linewidth]{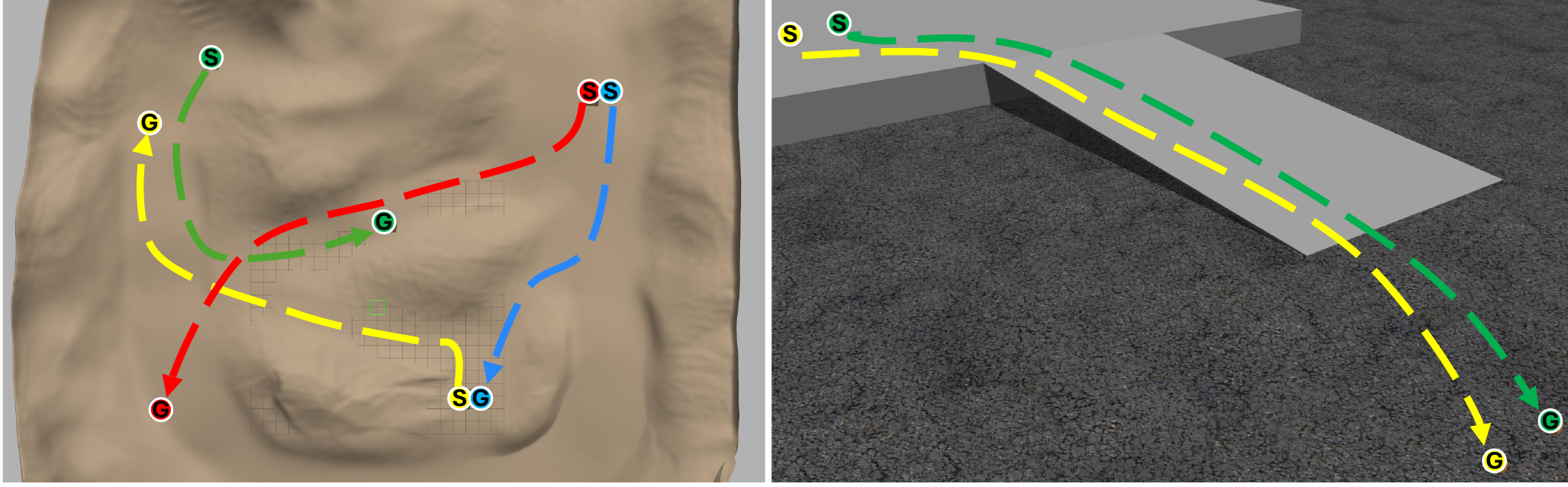}
    \caption{Simulation environments with start-goal pairs and expected path in Gazebo. Left: the mountainous terrain provided by the baseline. Right: the custom-built cliff terrain.}
    \label{fig:gazebo_sim}
    \vspace{-1.5em}
\end{figure}

We consider three navigation configurations to isolate the contributions of the terrain representation and the planner. The baselines are selected to provide a direct comparison with the existing mean-first approach and to disentangle the effects of the traversability representation from those of the planning algorithm.
(i) \textbf{Base-RRT*}, a baseline configuration that combines mean-first traversability cost with the RRT* planner \cite{leininger2024gaussian};
(ii) \textbf{Base-MPPI}, an ablation configuration that uses the same traversability representation as Base-RRT* while replacing RRT* with MPPI, thereby isolating the effect of the planner; 
and (iii) \textbf{Prob-MPPI}, the proposed configuration that integrates the probabilistic traversability representation with MPPI.

Furthermore, the influence of uncertainty sensitivity is evaluated by varying the coefficient $\factorconf_\delta$ in the uncertainty-aware terrain cost, with $\factorconf_\delta=1.645$ corresponding to a one-sided confidence level of $95\%$, $\factorconf_\delta=0.842$  to $80\%$, and $\factorconf_\delta=0$ reducing the terrain cost to the expected traversability cost and thereby disregards traversability uncertainty.
Note, this $\factorconf_\delta=0$ cost is still obtained from the proposed function-space sampling approach, where traversability is evaluated over sampled terrain realizations, rather than from a mean-first formulation that computes geometric features directly from the posterior mean elevation.

The experiments were conducted on a workstation equipped with Intel® Core™ Ultra 9 and NVIDIA RTX PRO 5000 Blackwell. 
All compared methods are evaluated under the same simulation environments, as shown in~\cref{fig:gazebo_sim}.
For the baseline, we use the officially released implementation without additional modifications or parameter tuning. 
For the proposed Prob-MPPI, the local terrain map covers a $10.2\times10.2$~m area with a spatial resolution of 0.2~m and is updated at 2~Hz.
The MPPI parameters are kept identical across all relevant configurations, with a control rate of 10~Hz, a prediction horizon of 80 steps, and 512 sampled trajectories per iteration, except for uncertainty coefficient $\factorconf_\delta$, which is varied only in the confidence-level study. 
The linear and angular velocities are constrained to $[0,1]$~m/s and $[-1.5,1.5]$~rad/s, respectively.

\begin{figure}[t]
    \centering
    \includegraphics[width=0.8\linewidth]{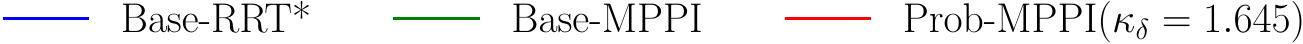}
    \includegraphics[width=1.0\linewidth]{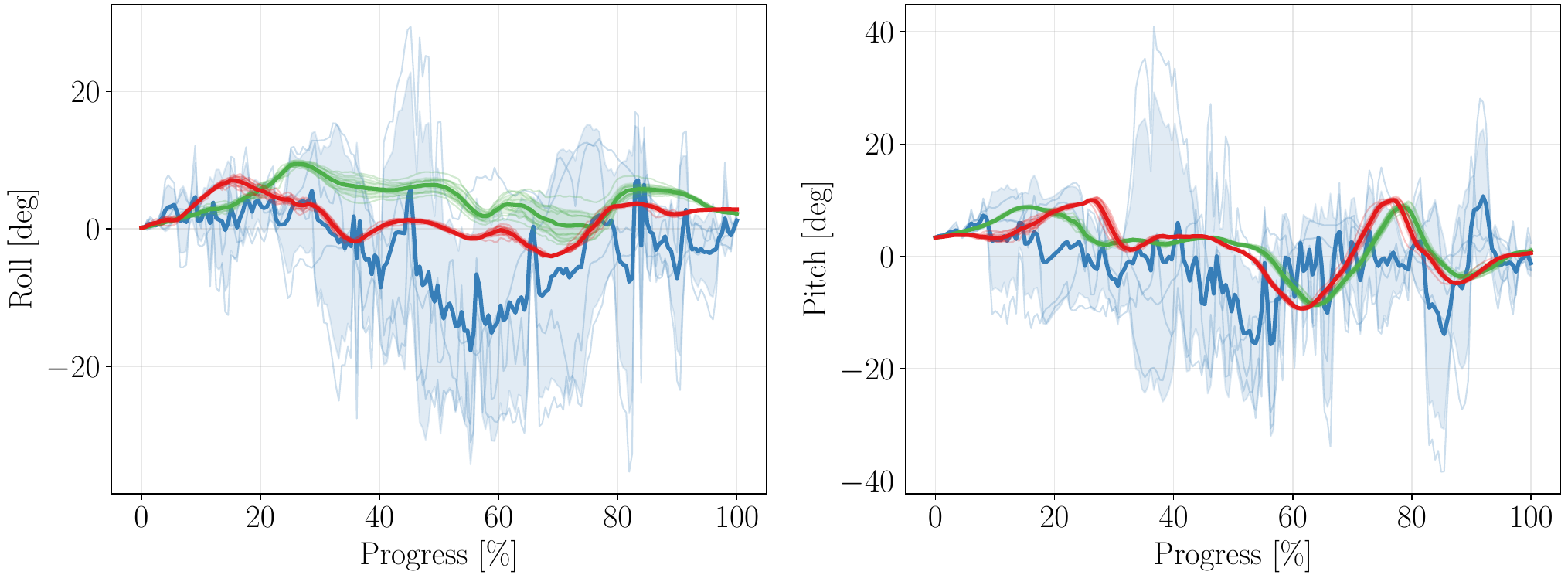}
    \caption{Robot attitude evolution during navigation. The robot navigates from the start position $(17, -17)\,$m to the goal $(-7, -7)\,$m in mountainous terrain. Only successful trials are shown.}
    \label{fig:sim_result_fig}
    \vspace{-1.5em}
\end{figure}
The \textbf{Mountain} comprises four start-goal scenarios on an uneven mountain terrain.
All configurations are evaluated in $10$ repetitions of each start-goal pair, focusing on body roll $\phi$ and pitch $\theta$ angles that characterize the robot attitude during terrain traversal and thus serve as indicators of its stability.
Besides, we report the navigation success rate as the proportion of trials where the robot successfully reaches the goal within a prescribed time limit without failure event such as rollover or falling from the terrain.
\cref{fig:sim_result_fig} compares the  roll and pitch responses recorded in an example scenario.
Both MPPI-based configurations exhibit generally smoother attitude responses than Base-RRT*, while Prob-MPPI produces the smallest roll and pitch excursions.
These results suggest that the probabilistic terrain representation provides MPPI with more informative terrain risk estimates, enabling the planner to avoid regions associated with unfavorable geometric conditions and thereby improving traversal stability.

The quantitative results in \cref{tab:sim_result_tab} indicate that on the mountain, all MPPI-based configurations achieve 100\% success across the tested start-goal pairs.
Replacing RRT* with MPPI substantially improves navigation reliability.
Moreover, Prob-MPPI exhibits lower mean and maximum roll and pitch deviations than Base-MPPI, even for $\factorconf_{\delta}=0$, where the uncertainty penalty is removed.
This suggests that the function-space sampling and probabilistic representation contribute to improve traversal stability by providing the planner with terrain costs that better reflect the underlying spatial structure of the terrain.
The three uncertainty levels result in similar attitude responses across the mountain scenarios. 
This can be attributed to the relatively dense observations and limited posterior uncertainty within the local planning horizon, making the uncertainty term small relative to the expected traversability cost and consequently limiting its influence on the resulting terrain cost.

\begin{table}[t]
\centering
\caption{%
Simulated Performance on Cliff Terrain, Measured by the Number of Falls and Success Rate over 10 Trials.
}
\label{tab:sim_result_cliff}
\resizebox{\linewidth}{!}{
\begin{tabular}{lll cc}
\toprule
\textbf{Start} (m) & \textbf{Goal} (m) & \textbf{Method} & 
\makecell{\textbf{No. of Falls}} & 
\makecell{\textbf{Success Rate}} \\
\midrule

\multirow{5}{*}{(-14, -1)} & \multirow{5}{*}{(7, -1)} 
   & Base-RRT*   & 5 & 10.00\% \\
 & & Base-MPPI   & 8 & 20.00\% \\
 & & Prob-MPPI($\factorconf_{\delta}=0$)   & 2 & 80.00\% \\
 & & Prob-MPPI($\factorconf_{\delta}=0.842$)  & 2 & 80.00\% \\
 & & Prob-MPPI($\factorconf_{\delta}=1.645$)  & \textbf{1} & \textbf{90.00\%} \\

\cmidrule(lr){1-5}

\multirow{5}{*}{(-14, -2)} & \multirow{5}{*}{(7, -2)} 
   & Base-RRT*  & 2 & 10.00\% \\
 & & Base-MPPI  & 10 & 0.00\% \\
 & & Prob-MPPI($\factorconf_{\delta}=0$) & 10 & 0.00\% \\
 & & Prob-MPPI($\factorconf_{\delta}=0.842$) & 8 & 20.00\% \\
 & & Prob-MPPI($\factorconf_{\delta}=1.645$)  & \textbf{1} & \textbf{90.00\%} \\

\bottomrule
\end{tabular}
}
\vspace{-1.25em} 
\end{table}
The \textbf{Cliff} comprises two start-goal scenarios over locally flat slopes with partially unobserved sections of negative obstacles. 
Since roll and pitch do not differ on such terrain, the quantitative results in \cref{tab:sim_result_cliff} focus on success rates only.
In the first scenarios, Base-RRT* and Base-MPPI achieve success rates of only 10\% and 20\%, respectively, while Prob-MPPI achieves 80-90\% depending on the uncertainty coefficient.
In the more challenging second scenario, the success rates range from 20\% for $\factorconf_{\delta}=0.842$ to 90\% for $\factorconf_{\delta}=1.645$.
By explicitly propagating terrain uncertainty into the traversability cost, Prob-MPPI assigns higher costs to regions with insufficiently reliable terrain estimates and consequently favors safer, better-observed regions further away from the cliff.
As a result, the robot avoids hazardous behaviors such as falling off the cliff and wheel lift-off. 
These results demonstrate the importance of uncertainty-aware traversability when terrain hazards cannot be reliably inferred from the available observations.

Notably, the same terrain cost weight is used for Prob-MPPI across all terrain types, without specific retuning. Despite the substantial difference between the mountain and cliff scenarios, the proposed method maintains high navigation reliability while adapting its behavior through the uncertainty-aware traversability cost. In contrast, the baseline methods exhibit substantially different performance across terrain types, indicating greater sensitivity to the terrain characteristics and planner tuning. This suggests that explicitly representing traversability uncertainty improves the navigation robustness to changes in terrain structure without requiring specific parameter adjustment.

\subsection{Deployment on a Mobile Robot}
The proposed approach has been deployed on a four-wheeled differential-drive robot equipped with a \texttt{Livox Mid-360} LiDAR and a \texttt{Xsens Vision Navigator} RTK-GNSS-IMU-camera fusion position unit.
The pipeline was configured to fit $4.2\times\SI{4.2}{\meter}$ elevation grid map with \SI{0.2}{\meter} cells from the point cloud data, performing GP hyper-parameter updates with $20$ epochs per map.
For MPPI planner, the prediction horizon is set to 80 time steps, with 512 sampled trajectories per iteration. The linear velocity is constrained to $[0, 1]$ m/s, while the angular velocity is bounded within $[-0.75, 0.75]$ rad/s.
The mapping and MPPI were running at \num{1} and \SI{5}{\hertz}, respectively, on an NVIDIA Jetson AGX Orin.

The system is deployed in three scenarios;
in the \textbf{Mound} scenario, the robot has to navigate to a goal position behind a mound of loose soil;
in the \textbf{Ramp} scenario, the robot drives downhill on surrounded by obstacles;
and in the \textbf{Sequence} scenario, the robot follows a set of waypoint to navigate approximately \SI{125}{\meter} through the environment, including driving downhill over the {Ramp}. 

\begin{figure}[t]
    \centering
    \includegraphics[width=0.99\columnwidth]{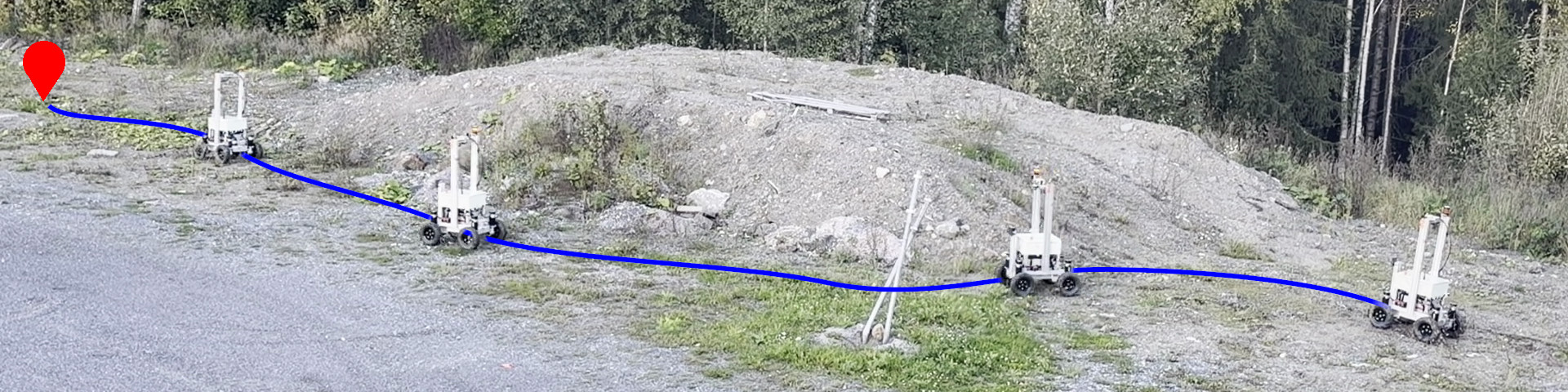}\\[0.1em]
    \phantom{}\hfill
    \includegraphics[width=0.4\columnwidth]{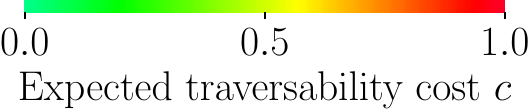}
    \hfill
    \includegraphics[width=0.4\columnwidth]{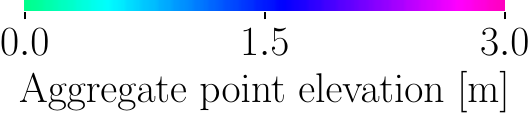}
    \hfill\phantom{}\\[0.1em]
    \includegraphics[width=0.99\columnwidth]{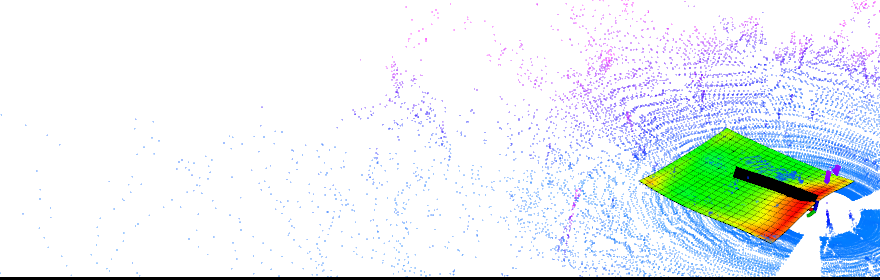}
    \includegraphics[width=0.99\columnwidth]{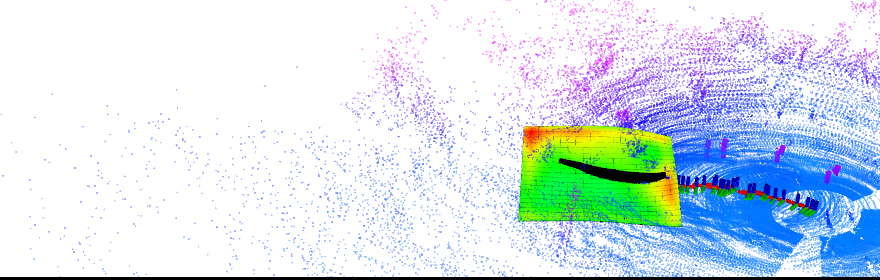}
    \includegraphics[width=0.99\columnwidth]{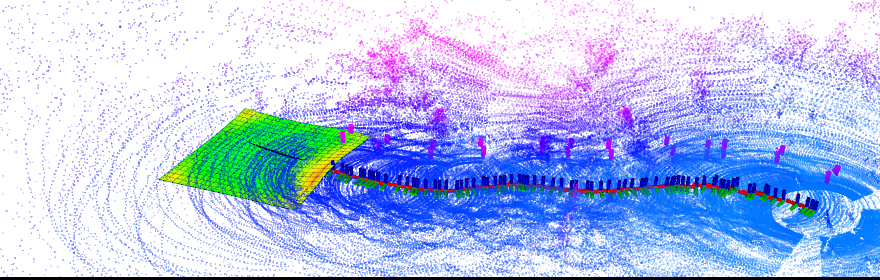}
    \includegraphics[width=0.99\columnwidth]{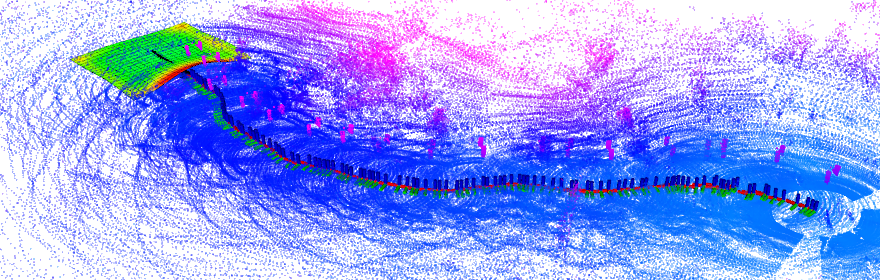}\\[-0.2em]
    \caption{%
       A typical behavior in the \textbf{Mound} scenario, where the robot is tasked to navigate to a goal behind a mound of loose soil.
       (top) A~view of the environment, with the robot position shown in approx. \SI{20}{\second} steps.
       The robot trajectory in blue, and its goal in red.
       (remaining rows) Visualization of the system at the time steps corresponding to the robot positions in the environment view;
       (1) the initial state facing the mound, (2) avoiding mound, (3) driving along the mound edge, and (4) approaching the goal. 
       The predicted cost grimap at the position (green-red), local MPPI planned path (black), and the trajectory so far.   
       Aggregate point cloud of the environment built from scans available so far in the execution (blue-purple); used only for visualization, while the robot navigates using the current scan.
    \label{fig:mound02fast}}
    \vspace{-1.5em}   
\end{figure}
In the \textbf{Mound} scenario, the robot starts facing the mound, with the shortest path to the goal available over the obstacle, but with an alternate flat area visible to one side, and a cliff to the other.
A typical execution is shown in~\cref{fig:mound02fast}, where the robot avoids the mound by driving over the flat ground.
Note that higher costs are reported in the robot's immediate vicinity, a consequence of using raw point cloud without close ground observations;
nonetheless, the system proves robust and capable of navigation using the raw point cloud.
The goal was reached in three repeated trials.
Besides, in one additional trial, the robot was erroneously initialized too close to the mound, resulting in a configuration where steering to the flat ground would navigate the robot over the mound's ramp side.
There, the system identified driving uphill as the best option, since the geometry is preferable to the mound's side.
However, the trial was stopped by the human supervisor, since the mound's loose soil provides only limited friction;
instead, we demonstrate the traversal over homogeneous slopes in the {Ramp} scenario next. 

\begin{figure}[t]
    \centering
    \includegraphics[width=0.99\columnwidth]{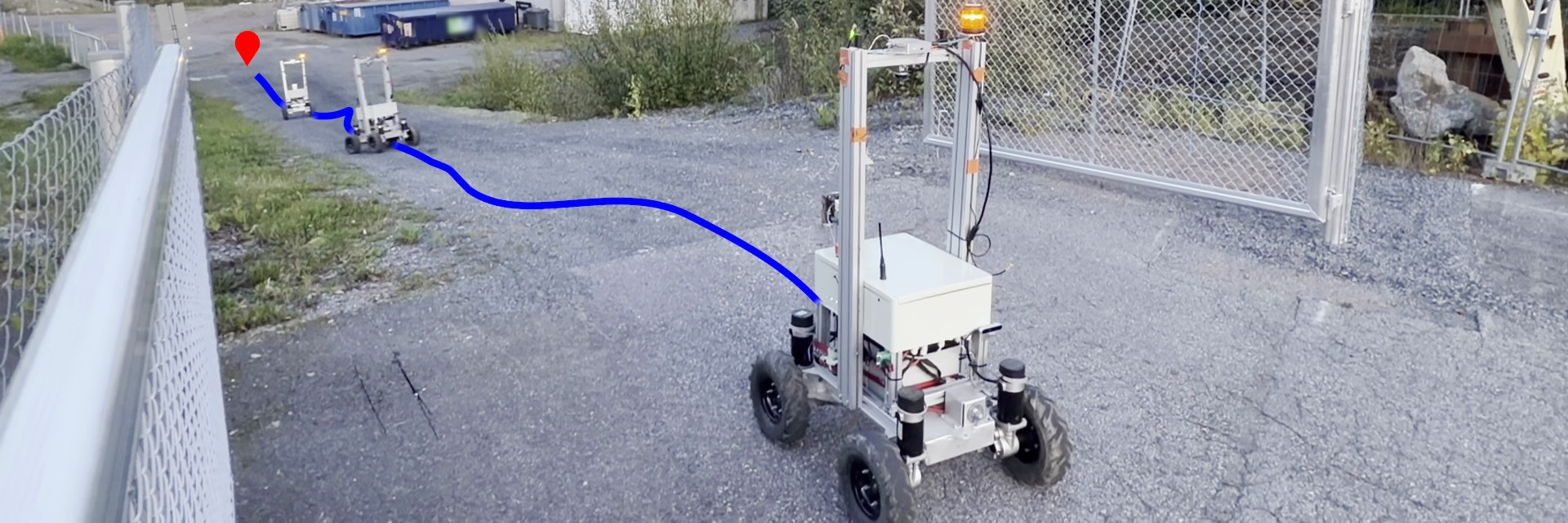}\\[0.1em]
    \phantom{}\hfill
    \includegraphics[width=0.4\columnwidth]{Figures/colorbar_cost.pdf}
    \hfill
    \includegraphics[width=0.4\columnwidth]{Figures/colorbar_pcdz.pdf}
    \hfill\phantom{}\\[0.1em]
    \includegraphics[width=0.325\columnwidth]{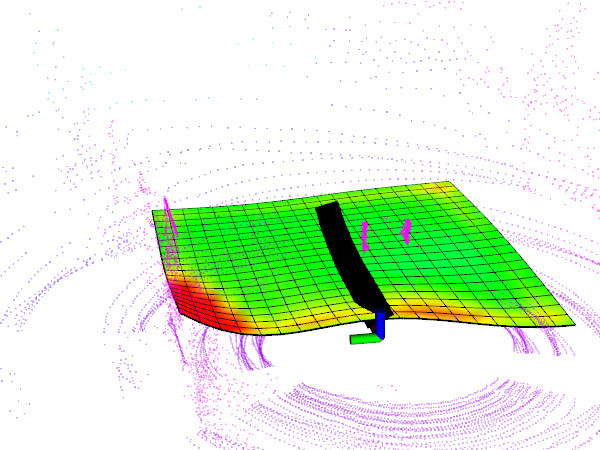}
    \hfill
    \includegraphics[width=0.325\columnwidth]{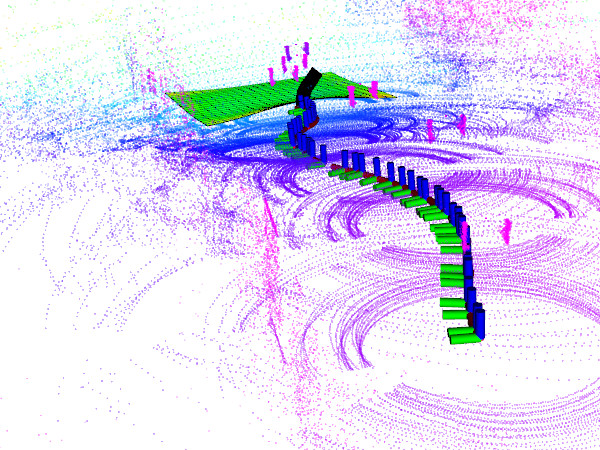}
    \hfill
    \includegraphics[width=0.325\columnwidth]{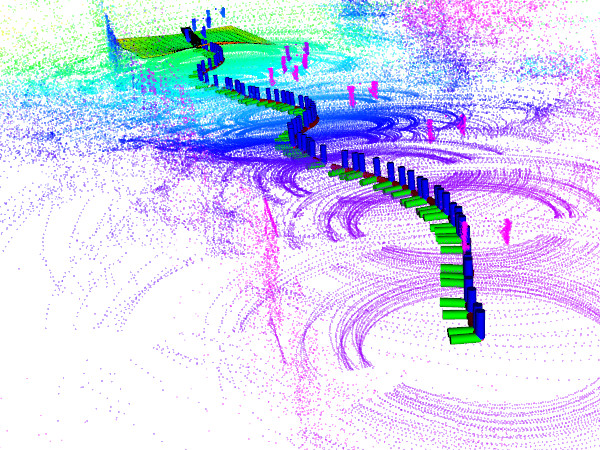}\\[-0.2em]
    \caption{%
       The \textbf{Ramp} scenario, where the robot is tasked to navigate downhill.
       (top) A view of the environment, with the robot position shown in approx. \SI{30}{\second} steps.
       The robot trajectory in blue, and its goal in red.
       (bottom) Visualization of the system at the time steps corresponding to the robot positions in the environment view;
       (1) the initial state uphill, (2) on the slope, and (3) approaching the downhill goal. 
       The predicted cost grid map at the position (green-red), local MPPI planned path (black), and the trajectory so far.   
       Aggregate point cloud built from scans available so far in the execution (blue-purple); used only for visualization, while the robot navigates using the current scan.
       }
       \vspace{-0.5em}  
    \label{fig:ramp}
\end{figure}
In the \textbf{Ramp} scenario, the system navigates towards the goal at the foot of hill as shown in~\cref{fig:ramp}.
Unlike in the {Mound} scenario, where the system can avoid the uneven ground, the {Ramp} must be traversed to reach the goal.
During operation, the system reports relatively low costs, since the slope is locally flat and thus only the step height $\steph$ and slope $\slope$ are agitated.

\begin{figure}[t]
    \centering
    \includegraphics[width=0.99\columnwidth]{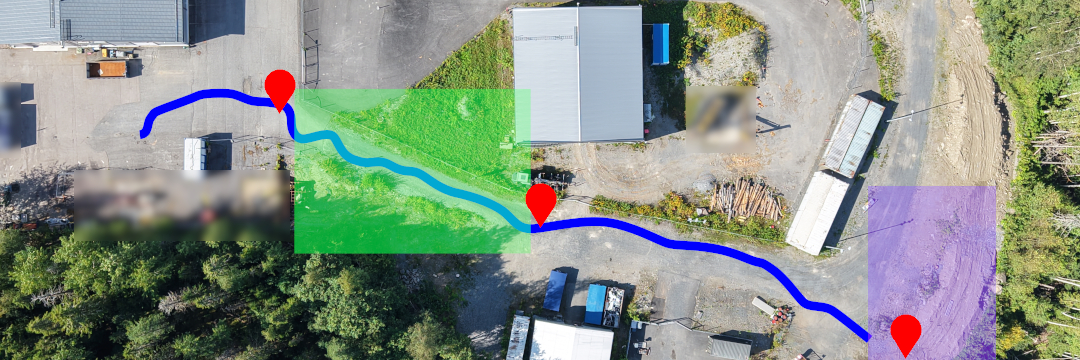}
    \vspace{-0.25em}   
    \caption{%
      The \textbf{Sequence} scenario.
      Overhead view of the test area, with the traversed path and waypoints of the {Sequence} scenario in blue and red, respectively, and the areas used in the {Mound} and {Ramp} scenario in purple and green, respectively.
    \label{fig:sequence}}
    \vspace{-1.25em}   
\end{figure}
Finally, the system is tasked to follow a \textbf{Sequence} of way-points as shown in \cref{fig:sequence}.
Overall, the mobile robot deployment demonstrates the suitability of ProTrav for onboard mobile robot navigation.
Nonetheless, the deployments also recall the inherent limitations of geometric approaches, which cannot distinguish additional visual contextual, such as discriminating the loose soil on the {Mound}.
There, we highlight that the proposed uncertainty propagation into traversability is largely feature and cost model agnostic, and could be applied to visual descriptors as well.
However, we leave such extension for future work.

\section{CONCLUSION}

We propose ProTrav, a probabilistic representation of terrain traversability for mobile robot navigation on uneven terrain.
A SVGP is used to construct a continuous elevation field, from which posterior function-space samples are generated using Monte Carlo sampling.
Terrain geometric features and traversability cost are evaluated over the individual terrain realizations, propagating uncertainty from elevation estimation to the resulting traversability cost. 
Then, the probabilistic representation is further integrated into an MPPI for risk-aware navigation.
The proposed approach is evaluated in simulation and real-world deployment.
Compared with conventional deterministic traversability baseline, explicitly representing traversability uncertainty helps the planner to better account for uncertain terrain geometry, improves attitude stability on uneven terrain, and enhances safety when facing negative obstacles. 

We suggest for the future work to focus both on the representation of fine geometric structures such as small pits that may be insufficiently captured from sparse observations, and on extensions to visual sensing.
In addition, because the probabilistic traversability representation is sensitive to observation changes, its estimated traversability may exhibit temporal fluctuations.
Future works should therefore investigate mechanisms for improving the temporal consistency and stability of the probabilistic representation.

\section*{Acknowledgements}
\textit{AI Tool Disclosure:}
Claude was used for code debugging.
ChatGPT and Gemini were used to create elements of~\cref{fig:system_design}.

\bibliographystyle{IEEEtran}
\bibliography{ref}

\end{document}